\documentclass{article}

\usepackage{microtype}
\usepackage{graphicx}
\usepackage{booktabs}
\usepackage{amsmath,amssymb}
\usepackage{xcolor}
\usepackage{multirow}
\usepackage{float}%
\usepackage[hidelinks]{hyperref}

\usepackage[accepted]{mlsys2025}
\makeatletter
\renewcommand{\mlsys@appearing}{Preprint.}
\makeatother
\hypersetup{pdfsubject={},pdftitle={Pando: Coupling Draft Trees to the Target's Sampling Noise for Edge Speculative Decoding}}

\newcommand{\istar}{I^{\ast}}%
\newcommand{\Nv}{N}%
\newcommand{\tacc}{\tau}%
\newcommand{\ldraft}{\ell_{\mathrm{draft}}}%
\newcommand{\lver}{\ell_{\mathrm{verify}}}
\newcommand{\ldec}{\ell_{\mathrm{decode}}}%
\newcommand{\edraft}{e_{\mathrm{draft}}}
\newcommand{\ever}{e_{\mathrm{verify}}}
\newcommand{\edec}{e_{\mathrm{decode}}}
\newcommand{\Fmax}{F_{\max}}
\newcommand{\Bmax}{B_{\max}}
\newcommand{\eop}{\epsilon_{\mathrm{op}}}
\newcommand{\ebyte}{\epsilon_{\mathrm{byte}}}
\newcommand{\Sgain}{S}%
\newcommand{\Ered}{R}%

\mlsystitlerunning{Pando: Noise-Coupled Parallel Draft Trees for Efficient Edge Speculative Decoding}

\begin{document}

\twocolumn[
	\mlsystitle{Pando: Noise-Coupled Parallel Draft Trees for Efficient Edge Speculative Decoding}

	\begin{mlsysauthorlist}
		\mlsysauthor{Joren Dumoulin}{kul}
		\mlsysauthor{Hyoukjun Kwon}{meta}
		\mlsysauthor{Subhankar Pal}{meta}        
		\mlsysauthor{Steven Li}{meta}
		\mlsysauthor{Vignesh Vivekraja}{meta}
		\mlsysauthor{Marian Verhelst}{kul}
	\end{mlsysauthorlist}
	\mlsysaffiliation{kul}{KU Leuven, Leuven, Belgium}
	\mlsysaffiliation{meta}{Meta}
	\mlsyscorrespondingauthor{Joren Dumoulin}{joren.dumoulin@kuleuven.be}
	\mlsyskeywords{speculative decoding, roofline, energy efficiency, edge inference}

	\vskip 0.3in
	\begin{abstract}

	Autoregressive (AR) decoding emits one token per forward pass, reloading the full
	model each time, so LLM inference on edge devices is memory-bound while the compute
	units sit idle. Speculative decoding puts that idle compute to work by verifying
	several drafted tokens in a single pass. State-of-the-art parallel drafters such as
	DFlash and DSpark propose an entire block of draft tokens in one forward pass, which
	makes drafting compute-efficient.
	However, they are deployed with draft chains whose
	length is capped by the block size they are trained for, leaving most of the
	computational budget for verification unused.

	To tackle this problem, we propose \emph{Pando}, which builds dynamic draft trees on top of a parallel
	drafter and thereby removes this cap. Pando employs \emph{noise-coupled pruning}, which draws
	the target model's sampling noise before drafting and ranks candidate branches under
	that noise, so the tree retains the branches the target will actually sample. Noise
	coupling is lossless and requires no retraining or additional model pass, yet
	increases acceptance length by up to $8.4\%$ on Qwen3-8B.
	To optimize and evaluate Pando configurations across hardware, we normalize an operator-level
	roofline to AR decoding, reducing every platform to its ridge point.
	On a representative edge device, a 16-node Pando tree attains $5.3\times$ speedup and
	$4.2\times$ energy reduction over AR decoding: $25\%$ faster than DSpark's released
	chain, and $27\%$ faster and $72\%$ more energy-efficient than an EAGLE-3 tree. The
	best configurations remain memory-bound on every edge device we model, so added
	compute capacity does not reduce their latency.

\end{abstract}

]
\printAffiliationsAndNotice{}

\section{Introduction}
\label{sec:intro}

Large language models (LLMs) are moving from the cloud to personal devices,
where applications such as on-device assistants must generate text under tight
latency, memory, and power constraints. The computational cost of LLM deployment
is dominated by the decode phase: autoregressive (AR) decoding produces one
token per forward pass, and each pass must stream the entire set of model
weights from memory. Decode is therefore memory-bound, and severely so on edge
devices, whose memory bandwidth is limited by package pins and power while their
matrix compute is not.

Speculative decoding~\cite{leviathan2023fast} breaks this one-token-per-pass
barrier. A lightweight drafter proposes several candidate tokens, and the target
model verifies all of them in a single forward pass using tree
attention~\cite{cai2024medusa}. If a round commits $\tau$ tokens on average, the
number of computationally expensive target forwards per token drops to $1/\tau$, turning idle
compute into decoding speed. The design question is how to allocate a
computational budget of $\Nv$ verified tokens to make $\tau$ as large as
possible.

\begin{figure}[t]\centering
            \includegraphics[width=\columnwidth]{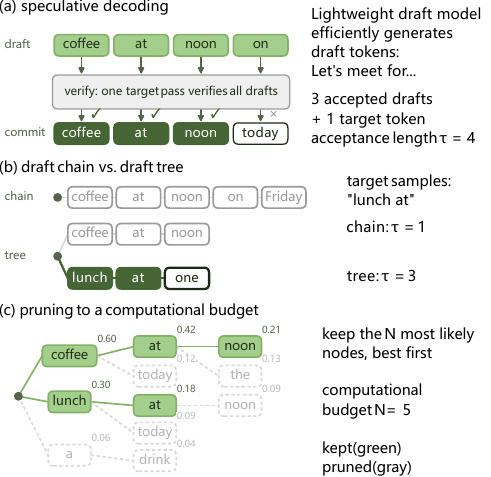}
      \vspace{-4mm}
      \caption{(a) Speculative decoding: a drafter proposes several tokens and the
          target verifies them in one pass, keeping the accepted prefix plus one
          token of its own. (b) With the same computational budget, a draft chain bets on
          one continuation, while a draft tree also covers alternatives the target
          may sample. (c) The tree in (b) is obtained by pruning a larger candidate
          tree to the computational budget, keeping the most likely nodes best first.}
    \vspace{-4mm}
      \label{fig:sd-overview}
  \end{figure}

Two families of drafters answer this question differently
(Fig.~\ref{fig:sd-overview}). Autoregressive drafters such as
EAGLE~\cite{li2024eagleb,li2025eagle} condition on the full prefix, so they can
produce a \emph{draft tree}---several alternative continuations---and prune it
dynamically to the computational budget, but a tree of depth $d$ requires $d$
drafter forwards. Parallel drafters, including the recent block
diffusion drafters DFlash~\cite{chen2026dflash} and
DSpark~\cite{cheng2026dspark}, instead emit an entire block of draft tokens with
one forward, which makes drafting compute-efficient and shifts almost all of the
computational cost of an iteration to verification. The catch is that a single
parallel forward returns one distribution per position, independent of which
tokens were drafted earlier, so it cannot score alternative branches. These
drafters are consequently evaluated only with \emph{chains}, capped at their
trained block length.

Parallel drafters were designed for high-concurrency serving, where concurrent
requests already consume the available compute and short chains suffice.
Single-stream edge decoding inverts this: with one request in flight, most of
the compute sits idle while the weights stream in, leaving a far larger
computational budget than a chain can use---and that a tree can. Limited battery and cooling also make
energy, not just latency, a first-class objective. The best computational
budget, the drafting method for using it, and how both depend on the device
remain open questions.

We study how to optimally deploy parallel drafters for single-stream edge
decoding, and make two observations.
First, parallel drafters \emph{can} build trees. DSpark's lightweight Markov
head rescores each candidate against the token preceding it, supplying exactly
the parent-conditioned probabilities that tree expansion requires. We use this
to build \textbf{Pando}, a dynamic draft tree for parallel drafters: one
backbone forward proposes per-position candidates, the head expands them level
by level, and the tree is pruned to the $\Nv$ most probable proposals. Pando
adds almost no drafting computational cost over a chain while lifting the 
block-length cap.

\begin{figure}[t]
    	\includegraphics[width=\linewidth]{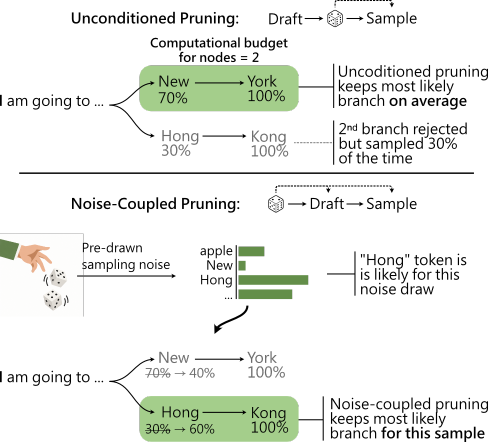}
	\caption{``New York'' and ``Hong Kong'' are both possible answers to the initial prompt. Unconditioned pruning always retains the ``New York'' branch, so the tree misses the target's sample 30\% of the time. Noise coupling prunes the tree conditioned on pre-drawn sampling noise and keeps the branch this draw favors.}
    \vspace{-4mm}
	\label{fig:noise-coupling}
\end{figure}

Second, and more fundamentally, conventional pruning is blind to how the target
samples. Current methods rank paths by cumulative draft probability, which
measures how likely a path is on average, independently of the target's
sampling. Under temperature sampling, however, verification descends only along
the token the target actually draws, and that token is frequently one of
moderate draft probability whose path a tree constrained to $\Nv$ nodes has
already pruned: the tree concentrates its nodes on the draft's modes and misses the
realized sample. To address this mismatch, we introduce \textbf{noise-coupled
pruning}. The Gumbel-max trick~\cite{maddison2014asampling} isolates all of the
target's sampling randomness into a noise vector that depends only on the
prefix, not on any model output, so it can be drawn \emph{before} drafting. We
pre-draw it and use it twice: to rescore and prune the tree, and to sample the
target (Fig.~\ref{fig:noise-coupling}). The tree is thus biased toward the
outcomes the target will actually sample, at essentially no computational
cost---no retraining, and no additional drafter or target forward. Because the committed token is always the target's own draw,
decoding remains lossless.

To compare configurations across devices rather than one platform, we map the
energy--latency design space across drafters, drafting methods, computational
budgets for verification, and platforms, combining measured acceptance with an
analytical computational cost model that counts the arithmetic operations and
off-chip bytes of every operator under an operator-level
roofline~\cite{williams2009roofline,hill2019gables,choi2013a,yuan2024llm}.
Normalizing by the computational cost of one AR decode step cancels absolute
peak compute and bandwidth, leaving a single platform parameter: the \emph{ridge
point}, the ratio of peak compute to peak bandwidth. Evaluating a configuration
then takes milliseconds, so we can cover the whole design space.

\begin{figure}[t]
    	\includegraphics[width=\columnwidth]{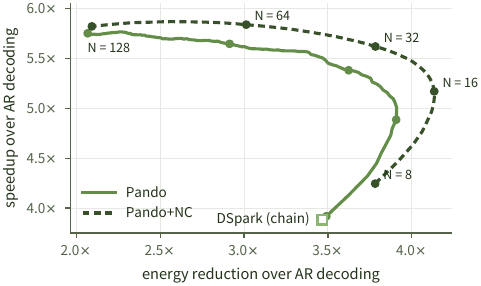}
    \vspace{-4mm}
	\caption{Speedup and energy reduction over AR decoding of DSpark with Pando, with and without noise coupling, at $T{=}1$ and computational budgets $\Nv$. Trees use larger computational budgets than DSpark's chain, and noise coupling improves both metrics further. Larger computational trade energy efficiency for speed.}
    \vspace{-4mm}
	\label{fig:pareto-teaser}
\end{figure}

We summarize our contributions as follows:
\begin{enumerate}\itemsep2pt
	\item \textbf{Pando}, a dynamic draft tree for parallel drafters, which lifts
	      their block-length cap at negligible computational cost.
	\item \textbf{Noise-coupled draft-tree pruning}: a lossless method that
	      biases the draft tree toward the target's sampled outcomes. It improves
	      acceptance length over plain tree pruning at every computational budget
	      we evaluate, most at small ones (\S\ref{sec:acceptance-nc}).
	\item \textbf{A design space map of tree-based speculative decoding} across
	      drafters, drafting methods, and computational budgets, which reduces
	      hardware to a single axis, the ridge point. It yields three insights.
	      Parallel drafters shift most computational work to the target's
	      verification, so optimizations that exploit drafter--target asymmetry
	      lose effect. The optimal computational budget grows with the ridge point
	      but saturates beyond roughly 64 FLOP/byte, below the 128--1024 FLOP/byte
	      of modern GPUs and NPUs, so the best configurations remain memory-bound
	      and extra compute for verification is futile; the optimal computational 
        budget is also independent of the target platform. Finally, the computational
        budget trades energy for latency: larger computational budgets favor speed,
        smaller ones energy efficiency (Fig.~\ref{fig:pareto-teaser}).
\end{enumerate}

\section{Background}
\label{sec:background}

\textbf{Speculative decoding.} Each iteration of speculative decoding has two steps. A drafter proposes
candidate tokens, and the target model verifies all of them in a single forward
pass. The target keeps the longest prefix of the proposal that it accepts and
then emits one more token of its own, the bonus token. We write $\Nv$ for the
number of tokens the target verifies per pass (the computational budget), and
$\tacc$ for the number of tokens committed per iteration,
including the bonus token. Autoregressive decoding has $\Nv=\tacc=1$.

The draft can be a chain or a tree. A chain is a single continuation, so
rejecting a token discards every token after it. A tree holds several
continuations: each node is a candidate token, its children are alternative next
tokens, and rejecting one child lets verification continue with a sibling. The
target verifies the whole tree in one pass by flattening it into a sequence and
masking attention so that each node sees only its ancestors
\cite{miao2024specinfer}. For a tree, $\Nv$ counts all nodes, including the
root.

Verification decides which draft tokens are accepted, while keeping the distribution of the output tokens equal to that of the target alone.
Let $p$ be the target distribution and $q$ the draft distribution. Rejection
sampling accepts a draft token $x$ with probability $\min(1,p(x)/q(x))$ and, on
rejection, samples a replacement from the residual $\max(0,p-q)$
\cite{leviathan2023fast,chen2023accelerati}. On a chain this yields the highest
possible acceptance probability per position, $\sum_x\min(p(x),q(x))$
\cite{sun2023spectr}. The rule is correct only if the draft tokens were sampled
from $q$, so trees verified this way must sample their children---with
replacement in SpecInfer \cite{miao2024specinfer}, without replacement in
Sequoia \cite{chen2024sequoia}. Sampled children can repeat a token or pick an
unlikely one, and their number per node must be fixed before sampling, so both
methods commit to a tree shape in advance: a fixed expansion per level in
SpecInfer, and one shape optimized offline per computational budget and
platform in Sequoia.

Target sampling removes the requirement that draft tokens are sampled, and this
is what makes dynamic trees possible. At each node the target samples its own
next token; if that token is one of the node's children, verification moves to
that child, and otherwise the iteration stops and the sampled token becomes the
bonus token. The output is the target's own sample whatever the tree contains,
so the drafter may choose both the tokens and the shape of the tree in every
iteration. Dynamic tree methods use this flexibility: they pick the most
probable children deterministically, grow the tree where the draft is confident,
and prune it to the $\Nv$ nodes with the highest cumulative draft probability
\cite{wang2025opttree,li2024eagle,xiong2024dyspec}. A fixed
computational budget $\Nv$ then goes to deep chains when the draft is
confident and to wide branching when it is not. This is the rule we use
for all trees. The ranking, however, uses only draft probabilities and ignores
which token the target will sample, so under sampling a small tree can spend its
nodes on likely paths that the sampled continuation does not take.

\textbf{Drafters.} Drafters differ in how many forward passes they need to produce a draft, and in
what distributions those passes return.

Autoregressive drafters produce one draft token per forward pass: either a
smaller model of the same family as the target
\cite{leviathan2023fast,chen2023accelerati}, or, in the EAGLE drafters, a small
autoregressive model run on the target's hidden features
\cite{li2024eagleb,li2024eagle,li2025eagle}. Conditioning on the full prefix
gives them a separate next-token distribution for every branch of a tree, at
a computational cost of $d$ passes for depth $d$.

Parallel drafters produce a whole block of draft tokens in one forward pass.
DFlash fills a block of placeholder positions in a single denoising step from
the target's hidden features \cite{chen2026dflash}; PARD and P-EAGLE obtain the
same one-pass structure with parallel draft heads
\cite{an2025pard,hui2026peagle}. The pass returns one distribution per position,
each independent of the token drafted at the previous position, so a parallel
drafter cannot distinguish branches and is used with chains. DSpark is a partial
exception: a small sequential Markov head on top of the parallel backbone
rescores each candidate given the token before it \cite{cheng2026dspark}, with
so little computational overhead that a block still requires about one pass. Even
so, DFlash and DSpark are evaluated only with chains, capped at the block size.
Parallel drafters thus make drafting compute-efficient, but offer no
established way to use a computational budget larger than their block.

\textbf{Hardware.} We target single-stream inference on edge devices. With only one request to
serve, every decode step loads the full model from memory to produce a single
token, and most of the compute sits idle while the weights stream in. How much
is left idle varies across devices, and the roofline model
\cite{williams2009roofline} captures it in a single number, the ridge point: the
ratio of a device's peak compute to its memory bandwidth. For devices built for
LLM inference it ranges from about 100 to 1000~FLOP/byte
(Appendix~\ref{app:devices}).

Speculative verification puts this idle compute to work. The target checks all
draft tokens in the same pass, reusing every weight it loads, so a larger tree
incurs little additional latency as long as the device has compute to spare. How
large a tree still provides benefits therefore depends on the device, yet existing work
fixes the computational budget for a single platform
\cite{chen2024sequoia,liu2026boosting}.
\section{Noise-Coupled Parallel Tree Drafting}
\label{sec:enabling}
\label{sec:noise-coupling}

\begin{figure*}[t]
	\centering
    	\includegraphics[width=\textwidth]{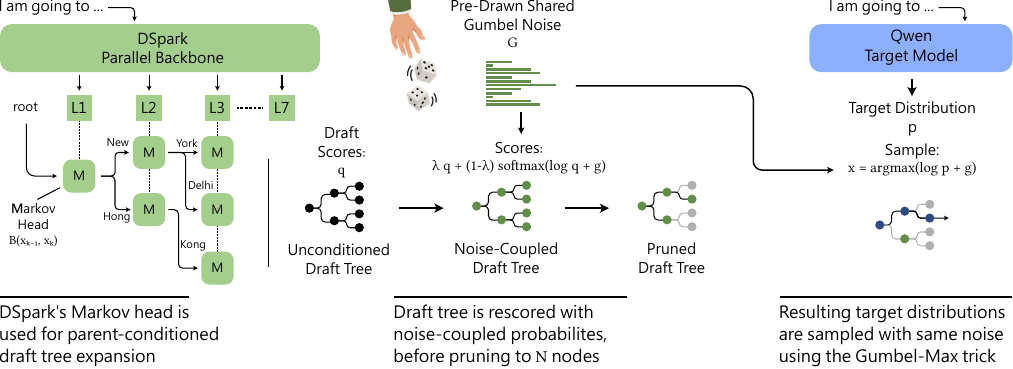}
    \vspace{-6mm}
	\caption{Noise-coupled tree drafting with DSpark. One backbone pass proposes
		per-position candidates, and the Markov head of the DSpark model rescores them against their
		parent to expand the draft tree. Pre-drawn Gumbel noise $g$ rescores the
		candidates before pruning to the $\Nv$ best nodes; the target samples with the
		same noise via the Gumbel-max trick.}
    \vspace{-4mm}
	\label{fig:nc-main}
\end{figure*}

DSpark proposes a whole draft block from one parallel backbone pass and refines
it with a lightweight Markov head, but its drafts are trained for fixed blocks,
so a computational budget beyond the block calls for a tree.
Throughout this section, $p$ is the target distribution, $q$ the draft distribution, and the
target samples at temperature $T$; noise-coupled results use $T=1$.

\subsection{Drafting under a computational budget of $\Nv$ nodes}
\label{sec:nc:baseline}

We follow the dynamic trees of EAGLE-2 and OPT-Tree
\cite{li2024eagle,wang2025opttree}, which build the tree in two
steps. The first step expands it. The root is the last committed token. Each
node is scored by the draft probability of its whole path from the root,
\begin{equation}
	s(x_{\le t}) = \prod_{j\le t} q(x_j\mid x_{<j}).
	\label{eq:cumprob}
\end{equation}
At each level, only the $k$ highest-scoring nodes of the previous level are
expanded.
The drafter provides a distribution over the next token for each of them, and each
receives its $k$ most probable tokens as children. After $d$ levels the tree
holds up to $1+k+(d-1)k^2$ nodes, usually more than the computational budget. The
second step
prunes it to the $\Nv$ nodes with the highest score. A path's score can only
drop as the path gets longer, so a node never scores above its parent. The kept
nodes therefore always form a tree: no node is kept without its ancestors. The
pruned tree is verified with target sampling (\S\ref{sec:background}).

This expansion needs a separate distribution for the children of every node, because
which token is likely next depends on the path that leads to it. An
autoregressive drafter provides this by running once per level on the
expanded nodes. A parallel drafter does not. DFlash, for instance, returns one
distribution $q_t(x_t)$ per position $t$, computed without knowing which tokens
were drafted at earlier positions. Every node at depth $t-1$ would then get the
same candidate children with the same probabilities, and the score of
Eq.~\eqref{eq:cumprob} could not tell whether consecutive tokens fit together.

DSpark restores the dependence on the previous token
\cite{cheng2026dspark}. Its parallel backbone still returns one distribution
$q_t(x_t)$ per position in a single pass. A small pretrained head $m$, a rank-256
Markov model over the vocabulary, then rescores each candidate given the token
directly before it. The draft probability of a child $x_t$ under a parent token
$x_{t-1}$ becomes
\begin{equation}
	q(x_t\mid x_{t-1}) \propto q_t(x_t)\,m(x_t\mid x_{t-1}),
	\label{eq:dspark-local}
\end{equation}
which differs between branches that end in different tokens. This is what tree
expansion needs. We run the backbone once and reuse its distributions for the
whole tree. Level by level, we pick the $k$ highest-scoring nodes of the previous level, apply the
head to each, give each its $k$ most probable children under Eq.~\eqref{eq:dspark-local}, and score
paths with Eq.~\eqref{eq:cumprob}, using Eq.~\eqref{eq:dspark-local} as the
per-token factor. We then prune to $\Nv$ exactly as in the reference
construction (Fig.~\ref{fig:nc-main}). The head runs once per level over the $k$
expanded nodes together, so the tree adds little drafting computational cost over a chain.
Because the head only sees the parent token, two branches that end in the same
token receive the same children. We call these DSpark draft trees \emph{Pando}.

\subsection{Pruning is blind to the sample}
\label{sec:nc:blind}

Under target sampling, verification descends the tree along the tokens the
target draws: at each depth the target samples a token, and the walk continues
only if that token is a child of the current node. Otherwise, the target's
distribution for the next token is unavailable, and a new draft--verify
iteration must start.
To use the target's sample during pruning, which precedes verification, we need
a sampling method whose randomness can be fixed before the target's distribution
is known.

The Gumbel-max trick \cite{maddison2014asampling} enables this: sampling from $\mathrm{softmax}(\log p/T)$ is
identical in distribution to perturbing every scaled log-probability with
independent Gumbel noise over the model's vocabulary, and taking the argmax, so the target's draw at output
position $j$ can be written as
\begin{equation}
      \begin{gathered}
          y_j = \arg\max_{x}\;\Bigl(\tfrac{1}{T}\log p(x\mid y_{<j}) + g_{y_{<j}}(x)\Bigr),\\
          g_{y_{<j}}(x)\sim\mathrm{Gumbel}(0,1),
      \end{gathered}
      \label{eq:gumbel-max}
\end{equation}
For example, let the target continue ``I am going to'' with ``New'' at
probability $0.7$ and ``Hong'' at $0.3$, at $T{=}1$. Their scores are
$\log 0.7 + g(\text{New}) = -0.36 + g(\text{New})$ and
$\log 0.3 + g(\text{Hong}) = -1.20 + g(\text{Hong})$. A noise draw
$g(\text{New}) = 0.1$, $g(\text{Hong}) = 1.3$ yields $-0.26$ against $0.10$, so the
target samples ``Hong''; over many draws this happens $30\%$ of the time, as
sampling from $p$ requires. Given the noise, the draw is a deterministic function of the target's probabilities: all of the sampling randomness sits in the noise vector $g_{y_{<j}}$, attached to the prefix $y_{<j}$, which does not depend on the model and can therefore be drawn before drafting. In the example, once this draw is known, a tree that keeps only the ``New'' branch is certain to miss. As $T\to0$, the draw reduces to the argmax of $p$ and the noise has no effect, so coupling the tree to the noise only matters for sampled decoding.

Ranking by cumulative draft probability does not use this information: it
allocates the computational budget to the draft's modes. At $T=1$ the target often
draws a token of moderate draft probability whose path is among the first
pruned, ending the iteration early even though the drafter assigned the token
non-negligible mass. Once drawn, however, the noise already favors some tokens
over others, and a tree that knows it can spend its computational budget on them. In
Fig.~\ref{fig:noise-coupling}, ``New'' is the more likely token on average, but this
noise draw favors ``Hong'', information that probability ranking leaves
unused.

\subsection{Noise-coupled scoring}
\label{sec:nc:method}

To couple drafting and verification, we draw the noise before drafting and use
it twice: to build the tree, and to sample the target with
Eq.~\eqref{eq:gumbel-max}. Each node $v$ carries one noise vector $g_v$ for the
position after it: the same vector the target will use if it samples at $v$. Its children share that vector, since they compete to be the
target's draw at $v$, while different prefixes get independent vectors.
For a candidate edge from $v$ to token $x$, DSpark's parent-conditioned
probability $q_x=q(x\mid v)$ is perturbed by the noise and mixed with its
unperturbed form,
\begin{equation}
	\tilde q_x = \lambda\, q_x + (1-\lambda)\,
	\mathrm{softmax}\Bigl(\tfrac{1}{T}\log q_x + g_v(x)\Bigr),
	\label{eq:nc-mixture}
\end{equation}
where the softmax runs over the expanded children of $v$ plus one aggregate
category for the remaining draft mass, which receives its own Gumbel draw. This
draw is a simplification: it is independent of the target's noise on the
unexpanded tokens, which cannot enter the tree but still compete for the target's
draw. A
path's score is the cumulative product of $\tilde q$ along it. As shown in Figure~\ref{fig:nc-main}, the
expanded tree is rescored this way and then pruned to the highest-scoring
ancestor-connected tree of $\Nv$ nodes, as in \S\ref{sec:nc:baseline}.

The second term in Eq.~\eqref{eq:nc-mixture} is a Gumbel-softmax sample
\cite{jang2017categorical,maddison2017concrete}: a smoothed form of the drafter's
own estimate of the target's draw, $\arg\max_x\bigl(\tfrac{1}{T}\log q_x +
	g_v(x)\bigr)$, which coincides with that draw when $q=p$. Unlike the one-hot
argmax, it keeps a graded ranking of the alternatives. Because the drafter is not
the target, its estimate can favor the wrong token; the $\lambda q_x$ term keeps
probability on the draft modes as a safeguard against that mismatch. $\lambda=1$
recovers cumulative-probability ranking. Acceptance peaks at $\lambda$ between
0.25 and 0.3 at every computational budget (Appendix~\ref{app:nc}); we use $\lambda=0.25$ for all computational budgets and call Pando with noise-coupled scoring \emph{Pando+NC}.

\subsection{Losslessness}
\label{sec:nc:lossless}

Noise coupling changes how fast the target generates text, not what it
generates. The output keeps exactly the target's distribution for any target
model and task, so a speedup needs no separate accuracy evaluation.

The reason is that the target makes the final decision. At each position, the
committed token is the target's own draw $y_j$ from Eq.~\eqref{eq:gumbel-max},
made with its own probabilities and the pre-drawn noise. The tree only decides
whether that token was proposed in advance, and therefore how many tokens one
pass can commit.

The drafter observes the noise, but only to form an estimate. The noise for a
position is drawn independently of everything before it, so knowing it does not change
which prefix led there. Whatever tree the drafter builds, $y_j$ remains an exact
sample from $\mathrm{softmax}(\log p(\cdot\mid y_{<j})/T)$.

This holds only if each position's noise is drawn once and then kept. If it were
drawn again in every iteration, a token that the tree happened to propose would
get additional opportunities to be committed, and the output would drift toward
the drafter's estimates. We therefore derive $g_v$ deterministically from the prefix
that ends at $v$, so the same position always sees the same noise.

\subsection{Computational overhead}
\label{sec:nc:cost}

Noise coupling is nearly computationally free. It retrains no drafter and adds no backbone or
target pass; the only new work is generating the noise, plus the mixture and
ranking, which reuse the drafter's existing outputs. Because the noise depends
only on the prefix, not on any model output, a counter-based generator keyed by
the prefix can produce exactly the entries needed, on demand: the noise of the
children the drafter expands, and the noise over the vocabulary at the nodes
where the target samples. The target's Gumbel-max sampling incurs approximately
the same computational cost as the standard sampling it replaces.

\section{Methodology}
\label{sec:method}

We rate each configuration by its speedup and energy reduction over
autoregressive (AR) decoding of the same target, at the same context and on the
same platform, using measured acceptance and modeled computational cost.
A configuration consists of a target and drafter pair, a drafting method, and a
computational budget $\Nv$ for verification. Chains are verified with rejection sampling and trees with target
sampling; for DSpark, the trees are Pando and Pando+NC (\S\ref{sec:enabling}).

\subsection{Speedup and energy reduction}
\label{sec:method:norm}
\label{sec:method:sep}

One speculative round has two sequential phases, drafting and verification. Let
$\ldraft$, $\lver$, and $\ldec$ be the latency of drafting, of verification,
and of one AR decode step, and $\edraft$, $\ever$, and $\edec$ the
corresponding energies. A round commits $\tacc$ tokens where an AR step commits
one, so the speedup $\Sgain$ and energy reduction $\Ered$ are

\begin{equation}
	\Sgain = \frac{\tacc \times \ldec}{\ldraft+\lver},\qquad
	\Ered = \frac{\tacc \times \edec}{\edraft+\ever}.
	\label{eq:norm}
\end{equation}

A larger tree raises $\tacc$ in the numerator but also the verification computational cost in
the denominator, differently for latency and energy. Since $\tacc$ depends only
on the models and the decoding protocol, not on the platform, we measure it on
a GPU and model latency and energy for each platform.

\subsection{Acceptance measurement}
\label{sec:method:acceptance}

We evaluate two drafters, EAGLE-3 and DSpark \cite{li2025eagle,cheng2026dspark}, with
Qwen3-8B as the target. We use the checkpoints of DeepSeek's DeepSpec release \cite{deepseek2026deepspec}, which trains both drafters in the
same setup for blocks of 7 draft tokens. All runs decode a single stream on nine
benchmarks (AIME25, Alpaca, Arena-Hard, GSM8K, HumanEval, LiveCodeBench, MATH500,
MBPP, MT-Bench), averaged with equal weight. Each drafter runs at $T{=}0$ and
$T{=}1$, with chains of 5 to 17 nodes and with trees of 4 to 128 nodes built as
in \S\ref{sec:nc:baseline}.

\begin{figure*}[tb]\centering
    	\includegraphics[width=\textwidth]{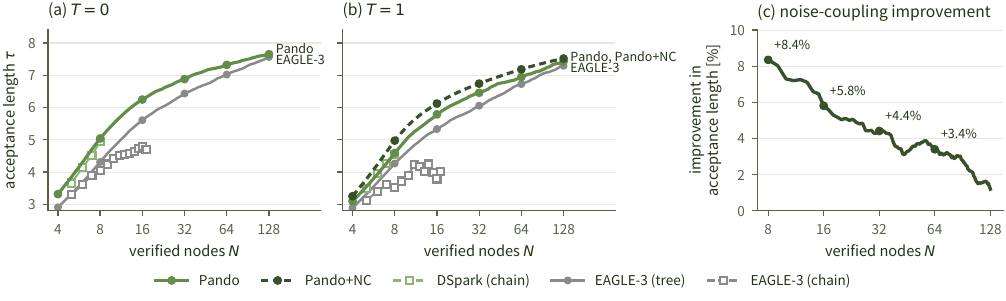}
    \vspace{-6mm}
	\caption{Acceptance length $\tacc$ against the computational budget for
		verification on Qwen3-8B, at $T{=}0$ (a) and $T{=}1$ (b). At $T{=}1$, Pando and Pando+NC are evaluated at every computational budget up to 128 nodes, with $\lambda{=}0.25$. (c) Relative improvement in acceptance length $\tacc$ from noise coupling.}
    \vspace{-4mm}
	\label{fig:acceptance}
\end{figure*}

\subsection{Computational cost model}
\label{sec:method:roofline}
\label{sec:method:energy}
The computational cost model rests on one observation: two devices with very different absolute speed can benefit equally from speculative decoding. The decode step and the verification pass both stream the full set of weights from memory, so a faster device speeds both up equally. What differs is how much extra arithmetic a pass can do before it waits for compute instead of memory, which is set by the ratio of peak compute to bandwidth, the ridge point. Expressed relative to an AR decode step, a platform's latency depends on that number alone.

We count, for every operator of the target and the drafter, the arithmetic
operations $F$ and the bytes $B$ moved to and from off-chip memory, for bf16
weights and activations and a 512-token context.

An operator takes as long as the slower of its compute and its memory traffic
\cite{williams2009roofline}:
\begin{equation}
	L = \max\!\left(\frac{F}{\Fmax},\frac{B}{\Bmax}\right)
	= \frac{B}{\Bmax}\max\!\left(\frac{I}{\istar},1\right),
	\label{eq:roofline}
\end{equation}
where $I=F/B$ is the operator's arithmetic intensity and $\istar=\Fmax/\Bmax$ the
ridge point of the platform. The maximum assumes that compute and memory traffic
overlap within an operator. The latency of a phase is the sum over its operators,
which assumes that operators run one after the other. To obtain speedup, we
normalize the latency of a phase $\mathrm{ph}\in\{\mathrm{draft},\mathrm{verify}\}$
by the latency of an autoregressive decode step, which yields
\begin{equation}
	\frac{\ell_{\mathrm{ph}}}{\ldec} = \frac{\sum_{i \in \mathrm{ph}} B_i \max(I_i/\istar,\,1)}
	{\sum_{j \in \mathrm{decode}} B_j \max(I_j/\istar,\,1)}.
	\label{eq:norm-roofline}
\end{equation}
Evaluating Eq.~\eqref{eq:norm-roofline} for $\mathrm{ph}=\mathrm{draft}$ and
$\mathrm{ph}=\mathrm{verify}$ yields the $\ldraft$ and $\lver$ terms of
Eq.~\eqref{eq:norm}. Every term includes the same factor $1/\Bmax$, so the peak
bandwidth and peak compute cancel from this equation.
What remains depends on the platform only through its ridge point $\istar$, the
ratio of its peak compute to its peak bandwidth.

Two devices with the same ridge point therefore yield the same speedup for
every configuration: the Apple~M5 and the RTX~5090 differ by more than
$10\times$ in peak compute but share a ridge point. Instead of modeling devices
one by one, we sweep a single axis, $\istar$.

Table~\ref{tab:devices} in Appendix~\ref{app:devices} places eleven devices targeting LLM inference on this axis. All of them lie between $100$ and $947$~FLOP/byte. Devices built for AI pack dense matrix units, tensor cores or NPUs,
next to memory whose bandwidth is limited by package pins and power, so their
compute grows much faster than their bandwidth. General-purpose CPUs have far
less matrix compute per byte: server parts such as the AMD EPYC~9654 and Intel
Xeon~Max sit around $35$ to $50$~FLOP/byte. Edge AI devices therefore offer far
more compute per byte than single-stream decoding uses, and this spare compute
is what speculative verification can use.
Although the roofline is coarse, the normalized verification computational cost is accurate.
It matches the detailed ZigZag-LLM model \cite{geens2024energy} within $6\%$ on a
cloud and an edge accelerator, and on an A100 it reproduces the measured
weight-bound plateau, with the measured computational cost 5 to $18\%$ higher between 16 and
128 nodes (Appendix~\ref{sec:ver-check}). Unlike these tools, it evaluates a
configuration in milliseconds, fast enough to sweep the whole design space.

For energy, every operation and every off-chip byte consumes a fixed amount of
energy \cite{choi2013a}:
\begin{equation}
	E = F\,\eop + B\,\ebyte .
	\label{eq:energy}
\end{equation}
Summing over a phase and dividing by the energy of a decode step yields
\begin{equation}
	e_{\mathrm{ph}} = \frac{F_{\mathrm{ph}}\,\rho+B_{\mathrm{ph}}}
	{F_{\mathrm{decode}}\,\rho+B_{\mathrm{decode}}},
	\qquad \rho = \eop/\ebyte.
	\label{eq:norm-energy}
\end{equation}
Energy reduces to a single number in the same way: a platform affects
$e_{\mathrm{ph}}$ only through the ratio $\rho$, so scaling both coefficients by the same factor, as a move to another process node roughly does,
leaves $e_{\mathrm{ph}}$ unchanged. We use $\ebyte=28$~pJ/byte for LPDDR5X
\cite{dally2020domain,nvidia2023grace} and $\eop=0.5$~pJ/op (2~TOPS/W) for bf16
\cite{horowitz2014computing}, and hold $\rho$ fixed across
the ridge-point sweep.

\section{Acceptance Length Across Computational Budgets}
\label{sec:acceptance}

\begin{figure*}[t]\centering
    	\includegraphics[width=\textwidth]{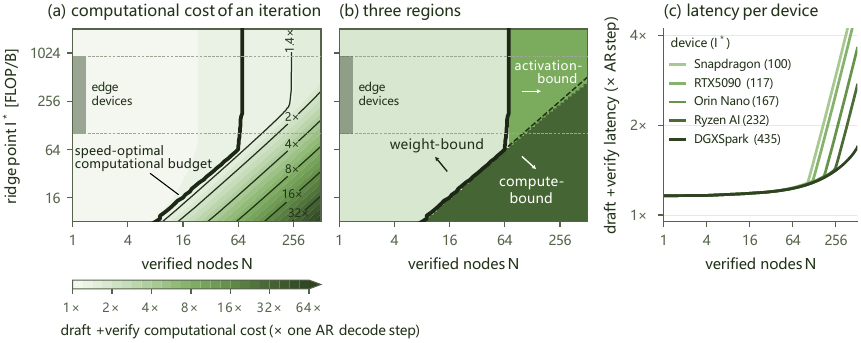}
    \vspace{-6mm}
	\caption{The computational cost of one Pando iteration as a function of two
		factors: (1) the number of nodes sent to verification and (2) the
		platform's ridge point. Plotted values present the combined drafting and
		verification computational cost, normalized to one autoregressive decode
		step for Qwen3-8B. (a) Computational cost contours. (b) The three regions
		in which execution is weight-bound, activation-bound, and compute-bound,
		separated by the speed-optimal computational budget. (c) Horizontal slices
		of (a) for five devices.
        }
    \vspace{-4mm}
	\label{fig:ver-vs-istar}
\end{figure*}

This section measures the numerator of Eq.~\eqref{eq:norm}: how the acceptance
length grows with the computational budget. Fig.~\ref{fig:acceptance} shows all
methods on Qwen3-8B, averaged over nine benchmarks; Appendix~\ref{app:bench}
lists them per benchmark.

\subsection{Acceptance against the computational budget}
\label{sec:acceptance-results}

Chains stop improving beyond the 8-node block the drafters are trained for, while
trees keep improving up to 128 nodes (Fig.~\ref{fig:acceptance}a,b); EAGLE-3's chain
levels off near $\tacc=4.7$ even when extended to 17 nodes. At 8 nodes, a tree improves little on a chain, $6.2\%$ for EAGLE-3 and
$2.0\%$ for Pando at $T{=}0$. Sampling hurts small trees further. A tree is accepted only if it contains the
target's own sample (\S\ref{sec:background}), and with 8 nodes ranked by draft
probability it often does not (\S\ref{sec:nc:blind}).

The tree's advantage is that it keeps improving with a larger computational
budget. At 128 nodes, $\tacc$ reaches $7.57$ for EAGLE-3 and $7.65$ for Pando at
$T{=}0$, $86\%$ and $55\%$ above the 8-node chains. The incremental improvement
diminishes: from 16 to 128 nodes, each doubling adds about $0.65$ to $\tacc$ for
EAGLE-3 and $0.47$ for Pando. EAGLE-3 conditions every level on the full prefix,
so deeper levels stay informative, but it requires a drafter pass per level. DSpark sees the branch only through the
parent token of its Markov head, so deeper levels add less.

\subsection{Noise-coupled pruning}
\label{sec:acceptance-nc}

Noise coupling raises acceptance at every computational budget, most at small
ones (Fig.~\ref{fig:acceptance}b,c). At $T{=}0$ the target's draw is the
argmax of $p$, so the noise has no effect.
The improvement (Fig.~\ref{fig:acceptance}c) is
largest at small computational budgets, where the tree can hold only a few
alternatives and choosing the ones this sample favors matters most. Live
decoding with Pando+NC confirms the improvement ($+7.0\%$ at 8 nodes and $+7.3\%$
at 16) with no measurable change in wall-clock time.

\section{Computational Cost of Iterations}
\label{sec:cost}
\label{sec:drafting}

\begin{figure*}[tb]\centering
			\includegraphics[width=\textwidth]{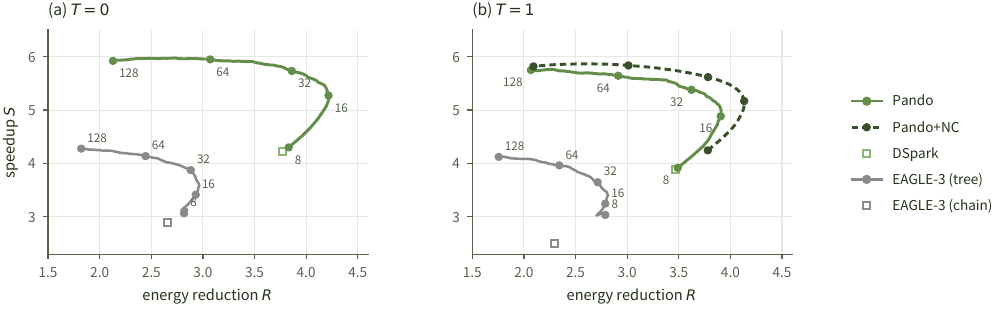}
	\vspace{-6mm}
	\caption{Speedup against energy reduction over AR decoding, Qwen3-8B, at
		$T{=}0$ (a) and $T{=}1$ (b). Trees trade energy for speed: a 16-node
		Pando tree is the most efficient, larger trees are faster but less
		efficient. Pando outperforms DSpark's chain and EAGLE-3 at every computational budget,
		and noise coupling (dashed) adds to both speed and efficiency under
		sampling.}
	\vspace{-4mm}

	\label{fig:pareto}
\end{figure*}

The computational cost of an iteration, the denominator of
Eq.~\eqref{eq:norm}, depends on only two quantities: the number of nodes $\Nv$
sent to verification and the platform's ridge point $\istar$. Drafting adds only
a small constant, because DSpark drafts a tree at almost the same computational cost
at any computational budget. Every Pando configuration on every device is therefore a point on one
map of the iteration computational cost relative to an AR decode step
(Fig.~\ref{fig:ver-vs-istar}).

\subsection{Three regions of the computational cost map}
\label{sec:verification}
\label{sec:ver-regions}
\label{sec:ver-slice}
\label{sec:ver-shapes}

Combining this computational cost with the acceptance lengths of \S\ref{sec:acceptance} gives
the speed-optimal computational budget as a function of the ridge point, the
line in Fig.~\ref{fig:ver-vs-istar}b. At low ridge points it is small: with
little spare compute, only a few nodes can be verified usefully. It rises with
the ridge point but saturates at about $64$~FLOP/byte, and divides the map into
three regions:

\emph{Weight-bound}: This is the regime of AR decoding, where compute is idle and
performance is dominated by LLM weight data transfers. The computational budget in this region is heavily underutilized, and larger draft trees
would improve speculative decoding speedup.

\emph{Compute-bound}: The added arithmetic of the extra nodes in the draft tree no
longer fits under the compute roof, and each node incurs latency. This increase in
latency is steep enough that the additional acceptance length of larger trees does 
not make up for the added computational cost.

\emph{Activation-bound}: Beyond about 64 nodes on platforms with a high ridge
point, the draft tree has grown considerably but still remains within the
computational capabilities of the platform. Each additional node now adds memory
traffic for its activations. While the model weights still dominate the total
traffic, the diminishing improvement from larger draft trees no longer justifies
this additional traffic.

\subsection{System implications}
\label{sec:ver-boundary}
\label{sec:ver-conclusion}
\label{sec:ver-energy}

With the optimal computational budget, Pando never becomes compute-bound on
typical edge devices. This has three implications.

\textbf{Compute beyond the computational budget is futile.} The speed-optimal computational
budget saturates at about 64 nodes above roughly $64$~FLOP/byte, where
acceptance grows too slowly to justify larger trees. A platform needs a ridge
point that matches the computational budget it runs, $\istar\approx\Nv$; beyond that, more
compute per byte adds no speed, and the optimal computational budget and speedup are the same
on every device.

\textbf{The best configurations are memory-bound.} The speed-optimal computational
budget, and the smaller most efficient one, lie in the weight-bound region on
every edge AI device. Adding compute, such as larger matrix units, does not speed
them up; only more bandwidth or less memory traffic, for example keeping
activations on chip, does.

\textbf{Drafting is no longer the bottleneck.} A parallel drafter is a small,
constant fraction of an iteration's computational cost, so accelerating or
offloading drafting helps little; the computational cost lies in the target's verification.

\section{End-to-End Speedup and Energy}
\label{sec:endtoend}

We now combine the measured acceptance with the modeled computational cost through
Eq.~\eqref{eq:norm} and ask what speculative decoding achieves end to end during the decode stage, in latency
and in energy, and what Pando and noise coupling add. The baseline is the way
DSpark is released: a chain of 7 draft tokens (8 nodes). We compare it with Pando
and Pando+NC, and include EAGLE-3 as an autoregressive drafter.
Fig.~\ref{fig:pareto} shows the speedup and energy reduction of every configuration.

\subsection{Speed and energy across computational budgets}
\label{sec:e2e-pareto}

Pando converts DSpark's idle computational budget into speed. Under greedy
decoding,
the released chain reaches $4.22\times$ speedup and $3.77\times$ energy reduction
over AR decoding. A 16-node Pando tree raises both, to $5.27\times$ and $4.22\times$
($+25\%$ and $+12\%$ over the chain), and a 64-node Pando tree reaches $5.95\times$
($+41\%$). Beyond 16 nodes, however, the energy reduction falls: at 64 nodes it
is $3.07\times$, and at 128 nodes only $2.13\times$, about half of the 16-node tree's (Fig.~\ref{fig:pareto}a).
Verification adds almost no latency up to 128 nodes (\S\ref{sec:cost}), but every
additional node still consumes arithmetic energy, while acceptance improves by
only about $0.5$ per doubling of the computational budget.

Speed and energy therefore favor different computational budgets. The fastest computational budget is about 64 nodes; the most
energy-efficient is about 16 nodes, which reduces speed by $11\%$ while improving
energy reduction by $37\%$.

At $T{=}1$, noise coupling moves every Pando tree further up and to the right
(Fig.~\ref{fig:pareto}b). The improvement is largest at the small computational
budgets that the energy optimum favors, where ranking by draft probability misses
the target's sample most often (\S\ref{sec:nc:blind}).
The 16-node Pando+NC tree reaches $5.17\times$ speedup and $4.13\times$ energy
reduction, $6\%$ above Pando on both and the most efficient configuration at
$T{=}1$. At larger computational budgets, where Pando already covers most likely
continuations, the improvement shrinks to about $1\%$ at 128 nodes.

\subsection{Comparison with an autoregressive drafter}
\label{sec:e2e-drafter}

Under greedy decoding, the EAGLE-3 tree at 128 nodes matches the acceptance length
of Pando, yet EAGLE-3 trails Pando on both speedup and energy reduction at every
computational budget and at both temperatures (Fig.~\ref{fig:pareto}). EAGLE-3
runs one drafter pass per tree level, so the depth that raises its acceptance also
raises its drafting computational cost, while DSpark drafts a tree at almost
the same small computational cost at any computational budget.

Compared as they are usually deployed, EAGLE-3 with a dynamic tree of about 64
nodes and DSpark with its released chain, the two drafters are close in speed:
under greedy decoding the chain is about as fast ($4.22\times$ against
$4.14\times$) but $54\%$ more energy-efficient. The parallel drafter's advantage
is then mainly efficiency. Pando's trees are what add speed: the
16-node Pando tree is $27\%$ faster than the EAGLE-3 tree and $72\%$ more
efficient, and the 64-node Pando tree is $44\%$ faster.
\section{Related Work}
\label{sec:related}

In the terms of Eq.~\eqref{eq:norm}, most work on speculative decoding raises the
acceptance length, while parallel drafters and hardware designs lower the
computational cost terms. Better drafters raise acceptance through autoregressive draft heads
\cite{li2024eagleb,li2025eagle,cai2024medusa,ankner2024hydra} or by drafting with
part of the target \cite{zhang2024draft,liu2024kangaroo,elhoushi2024layerskip}.
Parallel drafters lower the drafting computational cost by emitting a block per forward pass
\cite{an2025pard,hui2026peagle,xiao2024parallelsp,bhendawade2024speculativ,cheng2026dspark}
or by diffusion \cite{chen2026dflash,liu2025tidar}. Recent work builds draft trees
for them \cite{ringel2026accelerati,li2026from,zhou2026treespark}. Closest to
Pando, PCTree \cite{li2026from} and TreeSpark \cite{zhou2026treespark} also
expand DSpark trees with its Markov head. PCTree evaluates only greedy
decoding, and TreeSpark adapts the tree size to the load of a server with
concurrent requests. Neither couples the tree to the target's sampling noise,
and neither relates the computational budget to the device's ridge point or energy. Tree methods
decide how to allocate a computational budget over nodes
\cite{miao2024specinfer,li2024eagle,wang2025opttree,oh2026bastion,zhang2026costaware},
and verification rules raise acceptance for a fixed set of nodes
\cite{leviathan2023fast,chen2023accelerati,sun2023spectr,hu2025towards,sun2024block,weng2025traversal,bachmann2025judge,holsman2025fuzzy}.

Noise coupling is closest to work that shares randomness between drafter and
target. \citet{daliri2024coupling} couple the two samplers through shared noise
so that the output does not depend on the drafter, and multi-draft methods
\cite{sun2023spectr,hu2025towards} couple several drafts with the target through
optimal transport. These methods couple to verify a given draft, whereas we use
the target's noise earlier, to choose which branches the tree contains.

Our computational cost model builds on the roofline
\cite{williams2009roofline,ilic2014cacheaware,hill2019gables} and the LLM
computational cost models derived from it
\cite{yuan2024llm,isaev2023calculon,zhang2024llmcompass,agrawal2024vidur,sun2026hardware},
which we extend to tree verification and energy
\cite{choi2013a,ghane2018power,k2025pagoda,chung2025the,consortium2024mlperf,wang2026online,liu2026position}.
Selection of the computational budget has been treated as a computational cost
decision \cite{kong2026an,wang2026smart}, made hardware-aware for one platform
\cite{chen2024sequoia}, or derived from profiling
\cite{chordiya2026lossless,liu2025speculativ,dutta2026benchmarki,li2026configspec}.
Closest to our analysis, \citet{liu2026boosting} use a roofline-derived metric to
size a tree on one FPGA. We answer the same question as a function of the ridge
point, which covers any device.

Hardware and runtime designs each target one platform and one drafter: they place
drafting or verification in memory \cite{li2024specpim,he2025lpspec,zirui2026ahasd},
build dedicated accelerators \cite{dong2026a,ha2026smolpu,ha2026edgexpert}, split
drafter and target across processors
\cite{zhang2024dovetail,lv2025duodecodin,wei2025ghidorah,mesa2026compileras,li2025sled},
manage storage and power \cite{wang2026lever,yang2026pelm}, or speed up tree
verification kernels
\cite{zhong2025specmamba,wang2026bole,guan2025yggdrasil}.
With a parallel drafter, designs that accelerate drafting act on a computational
cost that is already small (\S\ref{sec:e2e-drafter}).

\section{Conclusion}
\label{sec:conclusion}

This paper presents noise-coupled pruning, a lossless method for constructing
draft trees that are tailored to the target model's sampling outcome. By drawing
the target's sampling noise before drafting and incorporating it into the
ranking of candidate branches, it retains the branches most likely to be
accepted, without retraining the drafter or adding model passes.
On Qwen3-8B, it increases the acceptance length by up to 8\%, with the
largest improvement at small computational budgets. Applied to
Pando, our tree-drafting extension of the DSpark parallel drafter, it further
improves both speedup and energy efficiency under sampling. To evaluate these
configurations across hardware platforms, we introduce a roofline-based
computational cost model normalized to autoregressive decoding, which
characterizes each platform by its ridge point alone. This analysis shows that a
16-node Pando tree is the most energy-efficient configuration on every edge
device we model, and that its verification remains memory-bound, so
additional compute capacity does not further reduce its latency.

\phantomsection\label{body:end}
\bibliographystyle{mlsys2025}
\bibliography{refs}

\onecolumn
\appendix

\section{Noise Coupling Details}
\label{app:nc}

\begin{figure}[!htb]\centering
	\includegraphics[width=0.9\textwidth]{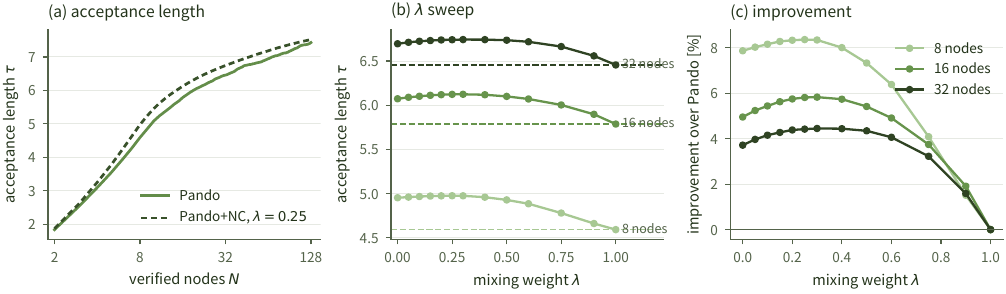}
	\caption{Noise coupling on DSpark, Qwen3-8B, $T{=}1$. (a) Acceptance
		length of Pando and Pando+NC ($\lambda{=}0.25$) against the
		computational budget. (b) Acceptance length against the mixing weight $\lambda$ at
		8, 16 and 32 nodes; $\lambda{=}0$ uses only the noise-perturbed distribution and
		$\lambda{=}1$ is Pando (dashed lines). (c) Improvement over Pando.}
	\label{fig:nc-detail}
\end{figure}

Fig.~\ref{fig:nc-detail} shows the acceptance length of Pando and Pando+NC and the
effect of the mixing weight. Acceptance peaks at $\lambda=0.25$ for 8 nodes and at $0.3$ for 16 and 32
nodes, within $0.03\%$ of $0.25$, and stays within $1\%$ of the peak for $\lambda$
between $0$ and $0.4$; it falls toward Pando at $\lambda=1$. We use $\lambda=0.25$
at every computational budget. We chose $\lambda$ on the evaluation
prompts, but because the curve is shallow around its peak, the choice inflates the
improvement little.

A live run on Qwen3-8B (six prompts, 64 tokens) confirms the improvement at
small computational budgets: the acceptance length of Pando+NC is $7.0\%$ above
Pando at 8 nodes and $7.3\%$ above at 16, with no measurable change in wall-clock
time. Its outputs can
differ from Pando's in bf16: the tree pass and the AR pass compute slightly
different logits, which can flip a near-tied Gumbel draw. In the cases we
audited with an fp32 target, the outputs match.

\section{Acceptance Length per Benchmark}
\label{app:bench}

Tables~\ref{tab:per-bench-t0} and~\ref{tab:per-bench-t1} list the acceptance length behind
Fig.~\ref{fig:acceptance} for each of the nine benchmarks. Acceptance varies strongly across tasks: math and code (AIME25, GSM8K, MATH500,
livecodebench) are the most predictable, open-ended chat (Alpaca, Arena-Hard, MT-Bench) the least.
Trees improve on every benchmark as the computational budget grows, and noise coupling raises
acceptance at small computational budgets on all but livecodebench.

\begin{table}[!htb]
	\centering\footnotesize
	\caption{Acceptance length $\tacc$ per benchmark on Qwen3-8B at $T{=}0$, for the 8-node chains and for trees at five computational budgets $\Nv$. The last column is the mean over benchmarks used in the main text; bold marks the highest value of each column. LCB is livecodebench.}
	\label{tab:per-bench-t0}
	\setlength{\tabcolsep}{3.6pt}
	\begin{tabular}{lrrrrrrrrrrr}
		\toprule
		Method & $\Nv$ & AIME25 & Alpaca & Arena-Hard & GSM8K & HumanEval & LCB & MATH500 & MBPP & MT-Bench & Mean \\
		\midrule
		DSpark chain & 8 & 5.74 & 2.52 & 3.17 & 6.48 & 5.89 & 6.80 & 6.22 & 4.73 & 2.94 & 4.94 \\
		\addlinespace
		Pando & 8 & 5.91 & 3.87 & 3.55 & 5.70 & 4.86 & 5.74 & 5.89 & 5.51 & 4.34 & 5.04 \\
		 & 16 & 7.80 & 4.45 & 3.95 & 7.40 & 5.96 & 6.84 & 7.91 & 6.72 & 5.20 & 6.25 \\
		 & 32 & 8.37 & 5.01 & 4.16 & 8.03 & 6.69 & 7.50 & 8.85 & 7.50 & 5.82 & 6.88 \\
		 & 64 & 8.66 & 5.49 & 4.42 & 8.36 & 7.23 & 8.01 & 9.25 & 8.19 & 6.29 & 7.32 \\
		 & 128 & 9.07 & \textbf{5.87} & \textbf{4.59} & 8.64 & \textbf{7.65} & 8.28 & \textbf{9.50} & \textbf{8.78} & \textbf{6.49} & \textbf{7.65} \\
		\addlinespace
		EAGLE-3 chain & 8 & 4.93 & 2.23 & 2.38 & 5.73 & 4.49 & 5.37 & 5.37 & 3.68 & 2.38 & 4.06 \\
		EAGLE-3 tree & 8 & 5.07 & 3.27 & 2.81 & 4.72 & 4.15 & 5.74 & 5.27 & 4.13 & 3.65 & 4.31 \\
		 & 16 & 7.26 & 3.90 & 3.44 & 6.25 & 5.35 & 7.58 & 7.17 & 5.21 & 4.31 & 5.61 \\
		 & 32 & 8.08 & 4.43 & 3.92 & 7.58 & 6.06 & 8.29 & 8.25 & 6.10 & 5.20 & 6.44 \\
		 & 64 & 9.14 & 4.85 & 4.25 & 8.26 & 6.71 & 8.87 & 8.70 & 6.71 & 5.70 & 7.02 \\
		 & 128 & \textbf{9.79} & 5.23 & 4.50 & \textbf{8.92} & 7.12 & \textbf{9.41} & 9.49 & 7.43 & 6.22 & 7.57 \\
		\bottomrule
	\end{tabular}
\end{table}

\begin{table}[!htb]
	\centering\footnotesize
	\caption{Acceptance length $\tacc$ per benchmark on Qwen3-8B at $T{=}1$, as in Table~\ref{tab:per-bench-t0}, including Pando+NC.}
	\label{tab:per-bench-t1}
	\setlength{\tabcolsep}{3.6pt}
	\begin{tabular}{lrrrrrrrrrrr}
		\toprule
		Method & $\Nv$ & AIME25 & Alpaca & Arena-Hard & GSM8K & HumanEval & LCB & MATH500 & MBPP & MT-Bench & Mean \\
		\midrule
		DSpark chain & 8 & 4.72 & 2.57 & 2.69 & 5.88 & 4.86 & 6.20 & 5.81 & 5.28 & 2.91 & 4.55 \\
		\addlinespace
		Pando & 8 & 4.76 & 3.72 & 3.18 & 5.22 & 3.79 & 6.16 & 5.74 & 5.06 & 3.72 & 4.59 \\
		 & 16 & 6.11 & 4.52 & 3.62 & 6.65 & 5.25 & 7.59 & 7.43 & 6.16 & 4.77 & 5.79 \\
		 & 32 & 6.99 & 5.20 & 3.74 & 7.30 & 6.02 & 8.05 & 8.18 & 7.08 & 5.55 & 6.46 \\
		 & 64 & 7.51 & 5.64 & 3.98 & 7.78 & 6.36 & 8.67 & 9.10 & 7.32 & 6.16 & 6.95 \\
		 & 128 & 7.83 & 5.99 & 4.19 & 8.49 & 6.98 & 8.98 & 9.70 & 8.19 & 6.51 & 7.43 \\
		\addlinespace
		Pando+NC & 8 & 5.12 & 4.02 & 3.32 & 5.87 & 4.58 & 6.20 & 6.04 & 5.34 & 4.28 & 4.98 \\
		 & 16 & 6.51 & 4.85 & 3.68 & 7.28 & 5.58 & 7.56 & 7.87 & 6.56 & 5.22 & 6.12 \\
		 & 32 & 7.29 & 5.39 & 3.90 & 8.04 & 6.17 & 8.09 & 8.65 & 7.34 & 5.81 & 6.74 \\
		 & 64 & 7.71 & 5.79 & 4.09 & 8.54 & 6.61 & 8.58 & 9.25 & 7.86 & 6.24 & 7.18 \\
		 & 128 & 7.96 & \textbf{6.05} & \textbf{4.27} & 8.90 & \textbf{7.00} & 8.97 & \textbf{9.71} & \textbf{8.26} & \textbf{6.54} & \textbf{7.52} \\
		\addlinespace
		EAGLE-3 chain & 8 & 3.74 & 1.79 & 2.10 & 4.47 & 3.84 & 4.62 & 5.21 & 3.69 & 2.20 & 3.52 \\
		EAGLE-3 tree & 8 & 4.68 & 3.29 & 2.64 & 4.41 & 3.58 & 6.83 & 4.96 & 4.31 & 3.73 & 4.27 \\
		 & 16 & 6.21 & 4.04 & 3.14 & 5.47 & 4.24 & 8.78 & 6.33 & 5.48 & 4.32 & 5.33 \\
		 & 32 & 6.72 & 4.40 & 3.55 & 7.04 & 4.91 & 9.34 & 7.33 & 6.43 & 4.82 & 6.06 \\
		 & 64 & 7.50 & 4.87 & 3.90 & 8.06 & 5.85 & 9.67 & 8.03 & 7.28 & 5.39 & 6.73 \\
		 & 128 & \textbf{8.04} & 5.15 & 4.25 & \textbf{9.13} & 6.25 & \textbf{10.37} & 8.58 & 7.63 & 6.23 & 7.29 \\
		\bottomrule
	\end{tabular}
\end{table}

\section{Devices and Computational Cost of Verification}
\label{app:devices}
\label{tab:map-read}

Table~\ref{tab:devices} lists the peak compute, memory bandwidth and ridge point
of the devices we use as reference points. Read at each device's ridge point, the
verification computational cost of Fig.~\ref{fig:ver-vs-istar} stays within
$1.02\times$ of a decode step at 16 nodes on every device. At
128 nodes, the seven devices above $160$~FLOP/byte stay within $1.13\times$, the
RTX~5090 and the Apple M5 chips reach $1.2\times$, and the Snapdragon
$1.4\times$.

\begin{table}[!htb]
	\centering\footnotesize
	\caption{Peak compute, memory bandwidth and ridge point of eleven well-known
		devices, sorted by $\istar$.}
	\label{tab:devices}
	\begin{tabular}{lrrr}
		\toprule
		Device                  & $\Fmax$   & $\Bmax$ & $\istar$ \\
		                        & [TFLOP/s] & [GB/s]  & [FLOP/B] \\
		\midrule
		Jetson AGX Thor         & 258.5     & 273.0   & 946.9 \\
		NVIDIA DGX Spark        & 118.8     & 273.0   & 435.2 \\
		NVIDIA H100 SXM         & 989.4     & 3350.0  & 295.3 \\
		Intel Core Ultra 9 288V & 33.6      & 136.5   & 246.2 \\
		Ryzen AI Max+ 395       & 59.4      & 256.0   & 232.0 \\
		Jetson AGX Orin         & 42.6      & 204.8   & 208.0 \\
		Jetson Orin Nano Super  & 17.0      & 102.0   & 166.7 \\
		NVIDIA RTX 5090         & 210.0     & 1792.0  & 117.2 \\
		Apple M5 (Metal)        & 17.9      & 153.0   & 117.1 \\
		Apple M5 Max (Metal)    & 71.7      & 614.0   & 116.7 \\
		Snapdragon 8 Gen 3      & 7.7       & 77.0    & 100.0 \\
		\bottomrule
	\end{tabular}
\end{table}

Context enters through the KV cache, which the tree mask reads once for all rows.
A longer context adds traffic that every row shares, so it moves the
activation-bound region to larger computational budgets. With unlimited compute,
the computational cost of a 128-node pass of Qwen3-8B falls from $1.12\times$ at 512 tokens to
$1.09\times$ at 32k.

\section{Accuracy of the Computational Cost Model}
\label{sec:ver-check}

\begin{figure}[!htb]\centering
	\includegraphics[width=0.6\textwidth]{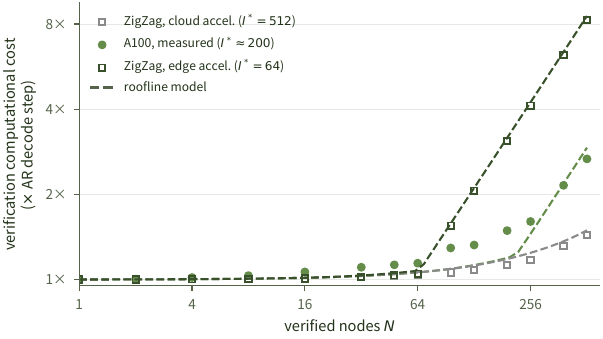}
	\caption{Qwen3-8B verification computational cost relative to one AR decode step,
		measured on an A100 (dots) and modeled by ZigZag-LLM for a cloud and an
		edge accelerator (squares), each against the roofline model at its ridge
		point (dashed).}
	\label{fig:ver-check}
\end{figure}

All computational costs in this paper are predictions from operator counts and peak
specifications. The roofline is the standard first-order model of LLM decoding
\cite{yuan2024llm,sun2026hardware}, and prior work on tree speculation sizes its
computational budget with a roofline-derived metric \cite{liu2026boosting}. To test it, we time
one verification pass of Qwen3-8B on an A100-SXM4-40GB at its default clock,
under the model's assumptions: bf16, a 512-token KV cache, $\Nv$ rows under a
tree mask, and the forward captured in a CUDA graph. Each point is the median of
40 passes. The model uses the measured ceilings of a large bf16 matrix product
and a streaming read, 275~TFLOP/s and 1377~GB/s ($\istar\approx200$~FLOP/byte).
As two further operating points, we model the same pass with ZigZag-LLM
\cite{geens2024energy}, which maps every layer onto the compute array and memory
hierarchy of an accelerator \cite{mei2021zigzag}. We use its cloud accelerator,
a $256\times256$ array with a 32~MB on-chip buffer ($\istar=512$~FLOP/byte), and
its edge accelerator, a $32\times32$ array with a 2~MB buffer
($\istar=64$~FLOP/byte). Fig.~\ref{fig:ver-check} compares all three, relative to
the AR step, with the roofline model.

On both ZigZag-LLM accelerators the roofline matches within $6\%$ at every
computational budget, and within $4\%$ up to 128 nodes, including the corner at
$\Nv\approx\istar$ on the edge accelerator; it is slightly high because it sends
all activations off chip, while the accelerators keep much of this traffic on chip. On the A100 the weight-bound plateau also holds: up to 64 nodes, the
pass stays within $1.15\times$ of a decode step. Between 16 and 128 nodes,
however, the measured computational cost is 5 to $18\%$ above the model, with a step between
64 and 96 nodes. The A100 is probably somewhat underutilized at these sizes: a
pass over 64 to 128 rows splits each weight matrix into fewer output tiles than
the GPU has streaming multiprocessors, so part of the chip idles while the
weights stream in, and the step marks where the matrix-multiply kernel switches
tile shape. Kernels tuned for such thin matrices, or an accelerator that maps
them well, should narrow this gap. At 512 nodes the model predicts $2.93\times$ and the
A100 takes $2.67\times$. All three operating points reproduce the weight-bound
plateau and where it ends, on which the conclusions of the paper rest.

\end{document}